\documentclass[11pt]{article}

\usepackage{acl2016}
\usepackage{times}
\usepackage{latexsym}
\usepackage{graphicx}
\usepackage{subcaption}
\usepackage{amsmath}
\usepackage{amsfonts}
\usepackage{mathrsfs}
\usepackage{bm}
\usepackage{booktabs}
\usepackage[para,online,flushleft]{threeparttable}
\usepackage{xcolor}
\usepackage{amsthm,paralist}
\usepackage{fancyhdr}
\aclfinalcopy 
\def\aclpaperid{804} 

\title{Compressing Neural Language Models by Sparse Word Representations}

\author{Yunchuan Chen,$^{1,2}$ Lili Mou,$^{1,3}$ Yan Xu,$^{1,3}$ Ge Li,$^{1,3}$ Zhi Jin$^{1,3,*}$\\
	$^1$Key Laboratory of High Confidence Software Technologies (Peking University), MoE, China\\
	$^2$University of Chinese Academy of Sciences, \texttt{chenyunchuan11@mails.ucas.ac.cn}\\
	$^3$Institute of Software, Peking University, \texttt{doublepower.mou@gmail.com},\\
	\texttt{\{xuyan14,lige,zhijin\}@pku.edu.cn}
	\quad $^*$Corresponding author
}

\date{}

\renewcommand{\vector}[1]{\bm{#1}}
\def\clap#1{\hbox to 2em{\hss#1\hss}}

\newcommand{\trans}[1]{#1^{\top}}
\def\Embedding{\texttt{Embedding}}
\def\Encoding{\texttt{Encoding}}
\def\Prediction{\texttt{Prediction}}

\DeclareGraphicsExtensions{.pdf,.jpg}

\begin{document}

\maketitle
\renewcommand{\headrulewidth}{0pt}
\cfoot{In \textit{Proc.~ACL}, pages 226--235, 2016.}
\thispagestyle{fancy}

\begin{abstract}
Neural networks are among the state-of-the-art techniques for language modeling. Existing neural language models typically map discrete words to distributed, dense vector representations. After information processing of the preceding context words by hidden layers, an output layer estimates the probability of the next word. Such approaches are time- and memory-intensive because of the large numbers of parameters for word embeddings and the output layer. In this paper, we propose to compress neural language models by sparse word representations. In the experiments, the number of parameters in our model increases very slowly with the growth of the vocabulary size, which is almost imperceptible. Moreover, our approach not only reduces the parameter space to a large extent, but also improves the performance in terms of the perplexity measure.\footnote{Code released on https://github.com/chenych11/lm}
\end{abstract}

\section{Introduction}\label{sec:intro}

Language models (LMs) play an important role in a variety of applications in natural language processing (NLP), including speech recognition and document recognition. In recent years, neural network-based LMs have achieved significant breakthroughs: they can model language more precisely than traditional $n$-gram statistics \cite{Mikolov:2011hq}; it is even possible to generate new sentences from a neural LM, benefiting various downstream tasks like machine translation, summarization, and dialogue systems \cite{Devlin:2014uh,rushneural,conversation,bf}.

Existing neural LMs typically map a discrete word to a distributed, real-valued vector representation (called \textit{embedding}) and use a neural model to predict the probability of each word in a sentence.
Such approaches necessitate a large number of parameters to represent the embeddings and the output layer's weights, which is unfavorable in many scenarios. First, with a wider application of neural networks in resource-restricted systems \cite{distill}, such approach is too memory-consuming and may fail to be deployed in mobile phones or embedded systems. Second, as each word is assigned with a dense vector---which is tuned by gradient-based methods---neural LMs are unlikely to learn meaningful representations for infrequent words. The reason is that infrequent words' gradient is only occasionally computed during training; thus their vector representations can hardly been tuned adequately.

In this paper, we propose a compressed neural language model where we can reduce the number of parameters to a large extent. To accomplish this, we first represent infrequent words' embeddings with frequent words' by sparse linear combinations. This is inspired by the observation that, in a dictionary, an unfamiliar word is typically defined by common words. We therefore propose an optimization objective to compute the sparse codes of infrequent words. The property of sparseness (only 4--8 values for each word) ensures the efficiency of our model.

Based on the pre-computed sparse codes, we design our compressed language model as follows.
A dense embedding is assigned to each common word;
an infrequent word, on the other hand, computes its vector representation by a sparse combination of common words' embeddings.
We use the long short term memory (LSTM)-based recurrent neural network (RNN) as the hidden layer of our model.
The weights of the output layer
are also compressed in a same way as embeddings.
Consequently, the number of trainable neural parameters is a constant regardless of the vocabulary size， if we ignore the biases of words. Even considering sparse codes (which are very small), we find the memory consumption grows imperceptibly with respect to the vocabulary.

We evaluate our LM on the Wikipedia corpus containing up to 1.6 billion words.
During training, we adopt noise-contrastive estimation (NCE) \cite{Gutmann:2012tr}
to estimate the parameters of our neural LMs.
However, different from \newcite{Mnih:2012tv},
we tailor the NCE method by adding a regression layer (called \texttt{ZRegressoion}) to predict the normalization factor,
which stabilizes the training process.
Experimental results show that, our compressed LM not only reduces the memory consumption, but also improves the performance in terms of the perplexity measure.

To sum up, the main contributions of this paper are three-fold.
(1) We propose an approach to represent uncommon words' embeddings by a sparse linear combination of common ones'.
(2) We propose a compressed neural language model based on the pre-computed sparse codes. The memory increases very slowly with the vocabulary size (4--8 values for each word).
(3) We further introduce a \texttt{ZRegression} mechanism to stabilize the NCE algorithm, which is potentially applicable to other LMs in general.

\section{Background}
\subsection{Standard Neural LMs}
Language modeling aims to minimize the joint probability of a corpus \cite{NLP}.
Traditional $n$-gram models impose a Markov assumption that a word is only dependent on previous $n-1$ words
and independent of its position.
When estimating the parameters, researchers have proposed various smoothing techniques including back-off models to alleviate the problem of data sparsity.

\newcite{NLM} propose to use a feed-forward neural network (FFNN) to replace the multinomial parameter estimation in $n$-gram models. Recurrent neural networks (RNNs) can also be used for language modeling; they are especially capable of capturing long range dependencies in sentences \cite{RNNLM,Sundermeyer:2015gj}.

\begin{figure}[!t]
\centering
\resizebox{0.25\textwidth}{!}{\includegraphics{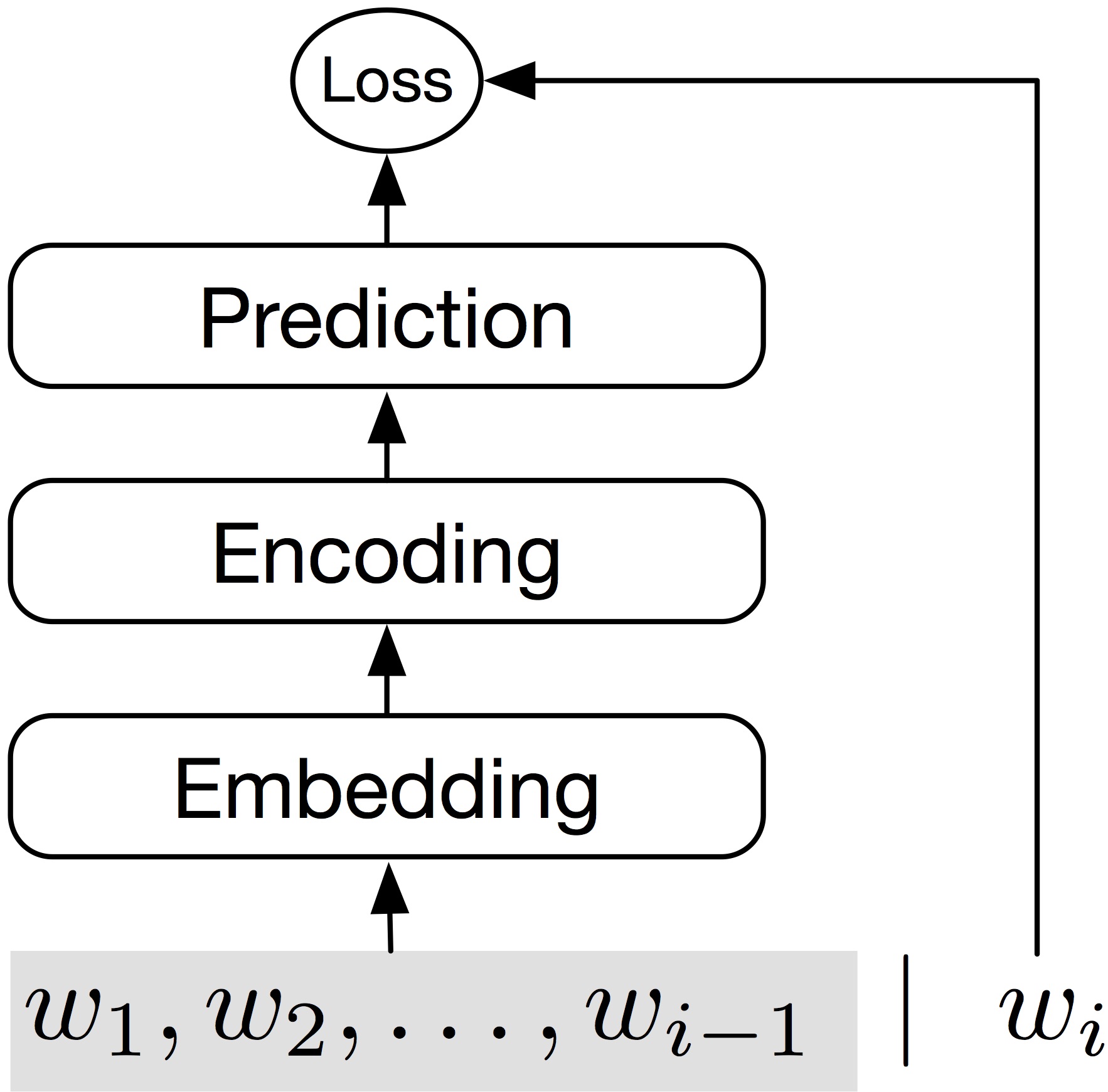}}
\caption{The architecture of a neural network-based language model.\label{fig:sNLM}}
\end{figure}

In the above models, we can view that a neural LM is composed of three main parts, namely the \Embedding{}, \Encoding{}, and \Prediction{} subnets, as shown in Figure~\ref{fig:sNLM}.

The \Embedding{} subnet maps a word to a dense vector, representing some abstract features of the word \cite{word2vec}. Note that this subnet usually accepts a list of words (known as history or context words) and outputs a sequence of word embeddings.

The \Encoding{} subnet encodes the history of a target word into a dense vector (known as \textit{context} or \textit{history} representation). We may either leverage FFNNs \cite{NLM} or RNNs \cite{RNNLM} as the \Encoding{} subnet, but RNNs typically yield a better performance \cite{Sundermeyer:2015gj}.

The \Prediction{} subnet outputs a distribution of target words as
\begin{align}
p(w = w_i|h) = &\dfrac{\exp(s(h, w_i))}
{\sum_j \exp(s(h, w_j))}, \label{eq:sNNLM}\\
s(h, w_i) = & \trans{\vector{W}_i} \vector{h} + b_i, \label{eq:sNNLM:score}
\end{align}
where $\vector{h}$ is the vector representation of context/history $h$, obtained by the
\Encoding{} subnet. $\vector{W} = (\vector{W}_1, \vector{W}_2, \dots, \vector{W}_V) \in \mathbb{R}^{C\times V}$ is the \textit{output weights} of \Prediction; $\vector{b} = (b_1, b_2, \dots, b_V) \in \mathbb{R}^{C}$ is the bias (the prior). $s(h, w_i)$ is a scoring function indicating the degree to which the context $h$ matches a target word $w_i$.
($V$ is the size of vocabulary $\mathcal{V}$; $C$ is the dimension
of context/history, given by the \Encoding{} subnet.)

\subsection{Complexity Concerns of Neural LMs}\label{sec:summ_sNNLM}

Neural network-based LMs can capture more precise semantics of natural language than $n$-gram models because the regularity of the \Embedding{} subnet extracts meaningful semantics of a word and the high capacity of \Encoding{} subnet enables complicated information processing.

Despite these, neural LMs also suffer from several disadvantages mainly out of complexity concerns.

\textit{Time complexity.} Training neural LMs is typically time-consuming especially when the vocabulary size is large. The normalization factor in Equation~(\ref{eq:sNNLM}) contributes most to time complexity.
\newcite{Morin:2005vo} propose hierarchical softmax by using a Bayesian network so that the probability is self-normalized.
Sampling techniques---for example, importance sampling
\cite{Bengio:2003tv}, noise-contrastive estimation
\cite{Gutmann:2012tr}, and target sampling \cite{Jean:NlGq_u6X}---are applied to avoid computation over the entire vocabulary.
Infrequent normalization maximizes
the unnormalized likelihood with a penalty term that favors normalized predictions \cite{Andreas:2015vm}.

\textit{Memory complexity and model complexity.} The number of parameters in the \Embedding{} and \Prediction{} subnets in neural LMs increases linearly with respect to the vocabulary size, which is large (Table~\ref{tab:no.param}).
As said in Section~\ref{sec:intro}, this is sometimes unfavorable in memory-restricted systems.
Even with sufficient hardware resources, it is problematic because we are unlikely to fully tune these parameters.
\newcite{Chen:2015wa} propose the differentiated softmax model by assigning fewer parameters to rare words than to frequent words.
However, their approach only handles the output weights, i.e., $\vector{W}$ in Equation~(\ref{eq:sNNLM:score}); the input embeddings remain uncompressed in their approach.

In this work, we mainly focus on memory and model complexity, i.e.,
we propose a novel method to compress the \Embedding{} and \Prediction{} subnets in neural language models.

\subsection{Related Work}
\textit{Existing work on model compression for neural networks.} \newcite{modelcompression} and \newcite{distill} use a well-trained large network to guide the training of a small network for model compression. \newcite{matrixfactor} compress neural models by matrix factorization, \newcite{quantization}  by quantization. In NLP, \newcite{distill2} learn an embedding subspace by supervised training. Our work resembles little, if any, to the above methods as we compress embeddings and output weights using sparse word representations. Existing model compression typically works with a compromise of performance. On the contrary, our model improves the perplexity measure after compression.

\textit{Sparse word representations. }
We leverage sparse codes of words to compress neural LMs.
\newcite{FaruquiTYDS15} propose a sparse coding method to represent each
word with a sparse vector. They solve an optimization problem to obtain the
sparse vectors of words as well as a dictionary matrix simultaneously.
By contrast, we do not estimate any dictionary matrix
when learning sparse codes,
which results in a simple and easy-to-optimize model.

\begin{table}

\centering
\small
		\begin{tabular}{lll}
			\toprule
			Sub-nets   & RNN-LSTM              & FFNN                   \\ \midrule
			{\small \Embedding{}}  & $VE$                  & $VE$                   \\
			{\small \Encoding{}}   & $4(CE+C^2+C)$         & $nCE +C$               \\
			{\small \Prediction{}} & $V(C+1)$              & $V(C+1)$               \\
			\textsc{total}$^\dag$  & $\mathcal{O}((C+E)V)$ & $\mathcal{O}((E+C)V )$ \\ \bottomrule
		\end{tabular}
		\caption{Number of parameters in different neural network-based LMs. $E$: embedding dimension; $C$: context dimension; $V$: vocabulary size. $^\dag$Note that $V \gg C \text{ (or $E$)}$.}
		\label{tab:no.param}
\end{table}

\section{Our Proposed Model}

In this section, we describe our compressed language model in detail.
Subsection~\ref{ss:sparse} formalizes the sparse representation of words, serving as the premise of our model.
On such a basis, we compress the \Embedding{} and \Prediction{} subnets in Subsections~\ref{ss:embedding} and~\ref{ss:prediction}, respectively.
Finally, Subsection~\ref{ss:NCE} introduces NCE for parameter estimation where we further propose the \texttt{ZRegression} mechanism to stabilize our model.

\subsection{Sparse Representations of Words}\label{ss:sparse}

We split the vocabulary $\mathcal{V}$ into two disjoint
subsets ($\mathcal{B}$ and $\mathcal{C}$). The first subset $\mathcal{B}$ is a base set, containing a fixed number of common words (8k in our experiments). $\mathcal{C}=\mathcal{V}\backslash \mathcal{B}$ is a set of uncommon words. We would like to use $\mathcal{B}$'s word embeddings to encode $\mathcal{C}$'s.

Our intuition is that oftentimes a word can be defined by a few other words, and that rare words should be defined by common ones. Therefore, it is reasonable to use a few common words' embeddings to represent that of a rare word. Following most work in the literature \cite{sparse,sparse2},
we represent each uncommon word with a sparse, linear combination of common ones' embeddings.
The sparse coefficients are called a \textit{sparse code} for a given word.

We first train a word representation model like SkipGram \cite{word2vec} 
to obtain a set of embeddings for each word in the vocabulary, including both common words and rare words. Suppose $\vector{U}=(\vector{U}_1, \vector{U}_2, \dots, \vector{U}_B)
\in \mathbb{R}^{E\times B}$ is the (learned) embedding matrix of common words,
i.e., $\vector{U}_i$ is the embedding of $i$-th word in $\mathcal{B}$.
(Here, $B=|\mathcal{B}|$.)

Each word in $\mathcal{B}$ has a natural sparse code (denoted as $\vector{x}$): it is a one-hot vector
with $B$ elements, the $i$-th dimension being on
for the $i$-th word in $\mathcal{B}$.

For a word $w\in \mathcal{C}$, we shall learn a sparse vector $\vector{x} = (x_1, x_2, \dots, x_B)$ as the sparse code of the word. Provided that $\vector{x}$ has been learned (which will be introduced shortly), the embedding of $w$ is
\begin{equation}\label{eq:sp2dense}
\hat{\vector{w}} = \sum_{j=1}^B x_j \vector{U}_j = \vector{U}\vector{x},
\end{equation}

To learn the sparse representation of a certain word $w$,
we propose the following optimization objective
\begin{align}
   \min_{\vector x}\ \ & \| \vector{Ux} - \vector{w} \|_{2}^{2} + \alpha \|\vector{x}\|_1 +
   \beta |\trans{\vector{1}} \vector{x} - 1| \nonumber\\
   &{} + \gamma \trans{\vector{1}} \max \{\vector{0}, -\vector{x}\},  \label{eq:opt_sp}
\end{align}
where $\max$ denotes the component-wise maximum;  $\vector{w}$ is the embedding for a rare word $w\in \mathcal{C}$.

The first term (called \textit{fitting loss} afterwards) evaluates the closeness between a word's coded vector representation and its ``true'' representation $\vector{w}$, which is the general goal of sparse coding.

The second term is an $\ell_1$ regularizer,
which encourages a sparse solution.
The last two regularization terms favor a solution that sums to 1 and
that is nonnegative, respectively.
The nonnegative regularizer is applied as in \newcite{document} due to psychological interpretation concerns.

It is difficult to determine the hyperparameters $\alpha$, $\beta$, and $\gamma$.
Therefore we perform several tricks. First, we drop the last term
in the problem \eqref{eq:opt_sp}, but clip each element in $\vector{x}$ so that all the sparse codes are nonnegative during each update of training.

Second, we re-parametrize  $\alpha$ and $\beta$ by balancing the fitting loss and regularization terms dynamically during training.
Concretely, we solve the following optimization problem, which is
slightly different but closely related to the conceptual objective \eqref{eq:opt_sp}:
\begin{equation}\label{eq:opt_sp_trick}
    \min_{\vector{x}}\ \ L(\vector{x}) + \alpha_t R_1(\vector{x}) + \beta_t R_2(\vector{x}),
\end{equation}
where $L(\vector{x}) = \| \vector{Ux} - \vector{w} \|_{2}^{2}$, $R_1(\vector{x}) = \|\vector{x}\|_1$,
and $R_2(\vector{x}) = |\trans{\vector{1}} \vector{x} - 1|$. $\alpha_t$ and $\beta_t$ are
adaptive parameters that are resolved during training time.
Suppose $\vector{x}_t$ is the value we obtain after the update of the $t$-th step,
we expect the importance of fitness and regularization remain unchanged during training.
This is equivalent to
\begin{align}
\dfrac{\alpha_t R_1(\vector{x}_t)}{L(\vector{x}_t)} = w_\alpha &\equiv \mbox{const}, \label{eq:alpha}\\
\dfrac{\beta_t R_2(\vector{x}_t)}{L(\vector{x}_t)} = w_\beta &\equiv \mbox{const}. \label{eq:beta}
\end{align}
or
\[
\alpha_t = \dfrac{L(\vector{x}_t)}{R_1(\vector{x}_t)} w_\alpha \text{ and }
\beta_t = \dfrac{L(\vector{x}_t)}{R_2(\vector{x}_t)} w_\beta,
\]
where $w_\alpha$ and $w_\beta$  are the ratios between the regularization loss and
the fitting loss. They are much easier to specify than $\alpha$ or $\beta$ in the problem~\eqref{eq:opt_sp}.

We have two remarks as follows.
\begin{compactitem}
\item To learn the sparse codes, we first train the ``true'' embeddings by {\tt word2vec}\footnote{https://code.google.com/archive/p/word2vec}  for both common words and rare words. However, these true embeddings are slacked during our language modeling.
\item As the codes are pre-computed and remain unchanged during language modeling, they are not tunable parameters of our neural model. Considering the learned sparse codes, we need only 4--8 values for each word on average, as the codes contain 0.05--0.1\% non-zero values, which are almost negligible.
\end{compactitem}

\subsection{Parameter Compression for the \Embedding{} Subnet}
\label{ss:embedding}
One main source of LM parameters is the \Embedding{} subnet, which takes a list of words (history/context) as input, and outputs dense, low-dimensional vector representations of the words.

We leverage the sparse representation of words mentioned above
to construct a compressed \Embedding{} subnet, where
the number of parameters is independent of the vocabulary size.

By solving the optimization problem \eqref{eq:opt_sp_trick} for each word, we obtain
a non-negative sparse code $\vector{x}\in\mathbb{R}^{B}$ for each word, indicating the degree to which the word is related to common words in $\mathcal{B}$.
Then the embedding of a word is given by $\hat{\vector{w}}=\vector{U}\vector{x}$.

We would like to point out that the embedding of a word $\hat{\vector{w}}$ is not sparse because $\vector{U}$ is a dense matrix, which serves as a shared parameter of learning all words' vector representations.

\subsection{Parameter Compression for the \Prediction{} Subnet} \label{ss:prediction}
Another main source of parameters is the \Prediction{} subnet.
As Table~\ref{eq:sNNLM} shows, the output layer contains $V$ target-word weight vectors and biases; the number increases with the vocabulary size.
To compress this part of a neural LM, we propose a weight-sharing method that uses words' sparse
representations again.
Similar to the compression of word embeddings, we define
a base set of weight vectors, and use them to represent the rest weights by sparse linear combinations.

Without loss of generality, we let $\vector{D} = \vector{W}_{:,1:B}$ be the output weights of $B$ base target words, and  $\vector{c} = \vector{b}_{1:B}$ be bias of the $B$ target words.\footnote{
$\vector{W}_{:,1:B}$ is the first $B$ columns of $\vector{W}$.} The goal is to use $\vector{D}$ and $\vector{c}$ to represent $\vector{W}$ and $\vector{b}$.
However, as the values of $\bm W$ and $\bm b$ are unknown before the training of LM, we cannot obtain their sparse codes in advance.

We claim that it is reasonable to share the same set of sparse codes to represent word vectors in \Embedding{} and the output weights in the \Prediction{} subnet. In a given corpus, an occurrence of a word is always companied by its context.
The co-occurrence statistics about a word or corresponding context are the same.
As both word embedding and context vectors capture these co-occurrence statistics
\cite{Levy:2014wb}, we can expect that context vectors share the same internal structure as embeddings.
Moreover, for a fine-trained network, given any word $w$ and its context $h$, the output layer's weight vector corresponding to $w$ should specify a large inner-product score for
the context $h$; thus these context vectors should approximate the
weight vector of $w$.
Therefore, word embeddings and the output weight vectors should share the same internal structures and  it is plausible to use a same set of sparse representations for both words and target-word weight vectors.
As we shall show in Section~\ref{sec:experiment}, our treatment of compressing the \Prediction{} subnet does make sense and achieves high performance.

Formally, the $i$-th output weight vector is estimated by
\begin{equation}
\hat{\vector{W}}_i = \vector{D} \vector{x}_i, \label{eq:sp2detector}
\end{equation}
The biases can also be compressed as
\begin{equation}
\hat{b}_i = \vector{c} \vector{x}_i. \label{eq:sp2detector_bias}
\end{equation}
where $\vector{x}_i$ is the sparse representation of the $i$-th word. (It is shared in the compression of weights and biases.)

In the above model, we have managed to compressed a language model whose number of parameters is
irrelevant to the vocabulary size.

To better estimate a ``prior'' distribution of words, we may alternatively assign an independent bias  to each word, i.e., $\vector{b}$ is not compressed. In this variant, the number of model parameters grows very slowly and is also negligible because each word needs only one extra parameter. Experimental results show that by not compressing the bias vector, we can even improve the performance while compressing LMs.

\subsection{Noise-Contrastive Estimation with \texttt{ZRegression}}\label{ss:NCE}
We adopt the noise-contrastive estimation (NCE) method to train our model.
Compared with the maximum likelihood estimation of softmax, NCE reduces computational complexity to a large degree.
We further propose the \texttt{ZRegression} mechanism to stablize training.

\begin{figure}
\begin{center}
	\resizebox{0.4\textwidth}{!}{
    	\includegraphics{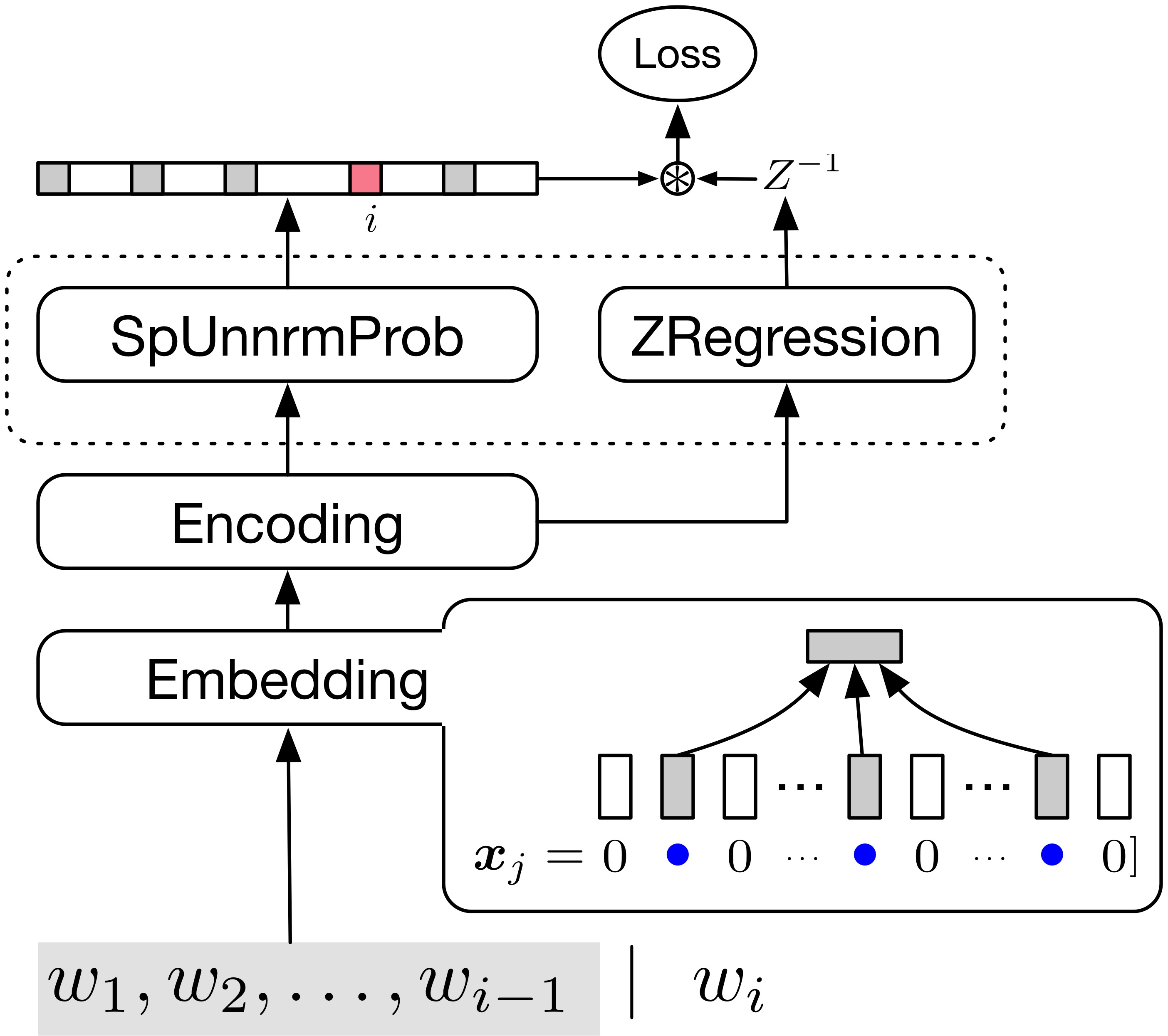}}
    \caption{\label{fig:NCE-regression} Compressing the output of neural LM. We apply NCE
    to estimate the parameters of the \Prediction{} sub-network (dashed round rectangle). The SpUnnrmProb layer outputs
    a \textbf{sp}arse, \textbf{unn}o\textbf{rm}alized \textbf{prob}ability of the next word. By ``sparsity,'' we mean that, in NCE, the probability is computed for only the ``true'' next word (red) and a few generated negative samples.}
\end{center}
\end{figure}

NCE generates a few negative samples for each positive data sample. During training, we only need to compute the unnormalized probability of these positive and negative samples.
Interested readers are referred to \cite{Gutmann:2012tr} for more information.

Formally, the estimated probability of the word $w_i$ with history/context $h$ is
\begin{align}
    P(w|h; \vector{\theta}) &= \dfrac{1}{Z_h} P^0(w_i|h; \vector{\theta})\nonumber \\
    &=
    \dfrac1{Z_h}\exp(s(w_i,h;\vector{\theta})),\label{eq:nrm_lang_model}
\end{align}
where $\bm\theta$ is the parameters and $Z_h$ is a context-dependent normalization factor.
 $P^0(w_i|h; \vector{\theta})$ is the unnormalized probability of the $w$
(given by the SpUnnrmProb layer in Figure~\ref{fig:NCE-regression}).

The NCE algorithm suggests to take $Z_h$
as parameters to optimize along with $\vector{\theta}$,
but it is intractable for context with variable lengths or large sizes in language modeling.
Following \newcite{Mnih:2012tv}, we set $Z_h=1$ for all $h$ in the base model (without \texttt{ZRegression}).

The objective for each occurrence of context/history $h$ is
\begin{align*}
J(\vector{\theta}|h) = & \log \dfrac{P(w_i|h;\vector{\theta})}
{P(w_i|h;\vector{\theta}) + k P_n(w_i)} + \\
  &   \sum_{j=1}^{k} \log \dfrac{kP_n(w_j)}
  {P(w_j|h;\vector{\theta}) + k P_n(w_j)},
\end{align*}
where $P_n(w)$ is the probability of drawing a negative sample $w$;
$k$ is the number of negative samples that we draw for each positive sample.

The overall objective of NCE is
\begin{equation*}
J(\vector{\theta}) = \mathbb{E}_h [J(\vector{\theta}|h)] \approx \dfrac{1}{M} \sum_{i=1}^{M} J(\vector{\theta}|h_i),
\end{equation*}
where $h_i$ is an occurrence of the context and $M$ is the total number of context occurrences.


Although setting $Z_h$ to 1 generally works well in our experiment, we find that in certain scenarios, the model is unstable. Experiments show that when the true normalization factor is far away from 1, the cost function may vibrate. To comply with NCE in general,
we therefore propose a \texttt{ZRegression} layer to predict the normalization constant $Z_h$ dependent on $h$, instead of treating it as a constant.

The regression layer is computed by
\[
Z_h^{-1} = \exp(\trans{\vector{W}_Z}\vector{h}+b_Z),
\]
where $\vector{W}_Z\in \mathbb{R}^{C}$ and $b_Z\in \mathbb{R}$
are weights and bias for \texttt{ZRegression}.
Hence, the estimated probability by NCE with \texttt{ZRegression} is given by
\begin{equation*}\nonumber
 P(w|h) = \exp(s(h,w)) \cdot \exp(\trans{\vector{W}_Z}\vector{h}+b_Z).
\end{equation*}

Note that the \texttt{ZRegression} layer does not guarantee normalized probabilities.
During validation and testing, we explicitly normalize the probabilities by Equation~\eqref{eq:sNNLM}.

\begin{table}[!t]
\centering
\resizebox{!}{!}{
\begin{tabular}{lr}
	\toprule
	Partitions         & Running words \\ \midrule
	Train ($n$-gram)   &         1.6 B \\
	Train (neural LMs) &         100 M \\
	Dev                &         100 K \\
	Test               &           5 M \\ \bottomrule
\end{tabular}
}
\caption{Statistics of our corpus.\label{tab:corpus_stat}}
\end{table}

\section{Evaluation}\label{sec:experiment}
In this part, we first describe our dataset in Subsection~\ref{ss:dataset}.
We evaluate our learned sparse codes of rare words in Subsection~\ref{ss:eval.code} and the compressed language model in Subsection~\ref{ss:eval.LM}. Subsection~\ref{ss:eval.zregression} provides in-depth analysis of the \texttt{ZRegression} mechanism.

\subsection{Dataset}\label{ss:dataset}

We used the freely available Wikipedia\footnote{http://en.wikipedia.org} dump (2014) as our dataset.
We extracted  plain sentences from the dump and removed all markups.
We further performed several steps of preprocessing
such as text normalization, sentence splitting, and tokenization.
Sentences were randomly shuffled,
so that no information across sentences could be used, i.e., we did not consider cached language models. The resulting corpus contains about 1.6 billion running words.

The corpus was split into three parts for training, validation, and testing.
As it is typically time-consuming to train neural networks,
we sampled a subset of 100 million running words to train neural LMs,
but the full training set was used to train the backoff $n$-gram models.
We chose hyperparameters by the validation set and reported model performance on the test set.
Table~\ref{tab:corpus_stat} presents some statistics of our dataset.

\subsection{Qualitative Analysis of Sparse Codes}\label{ss:eval.code}

\begin{figure}[!t]
\centering
	\resizebox{0.48\textwidth}{!}{
		\includegraphics{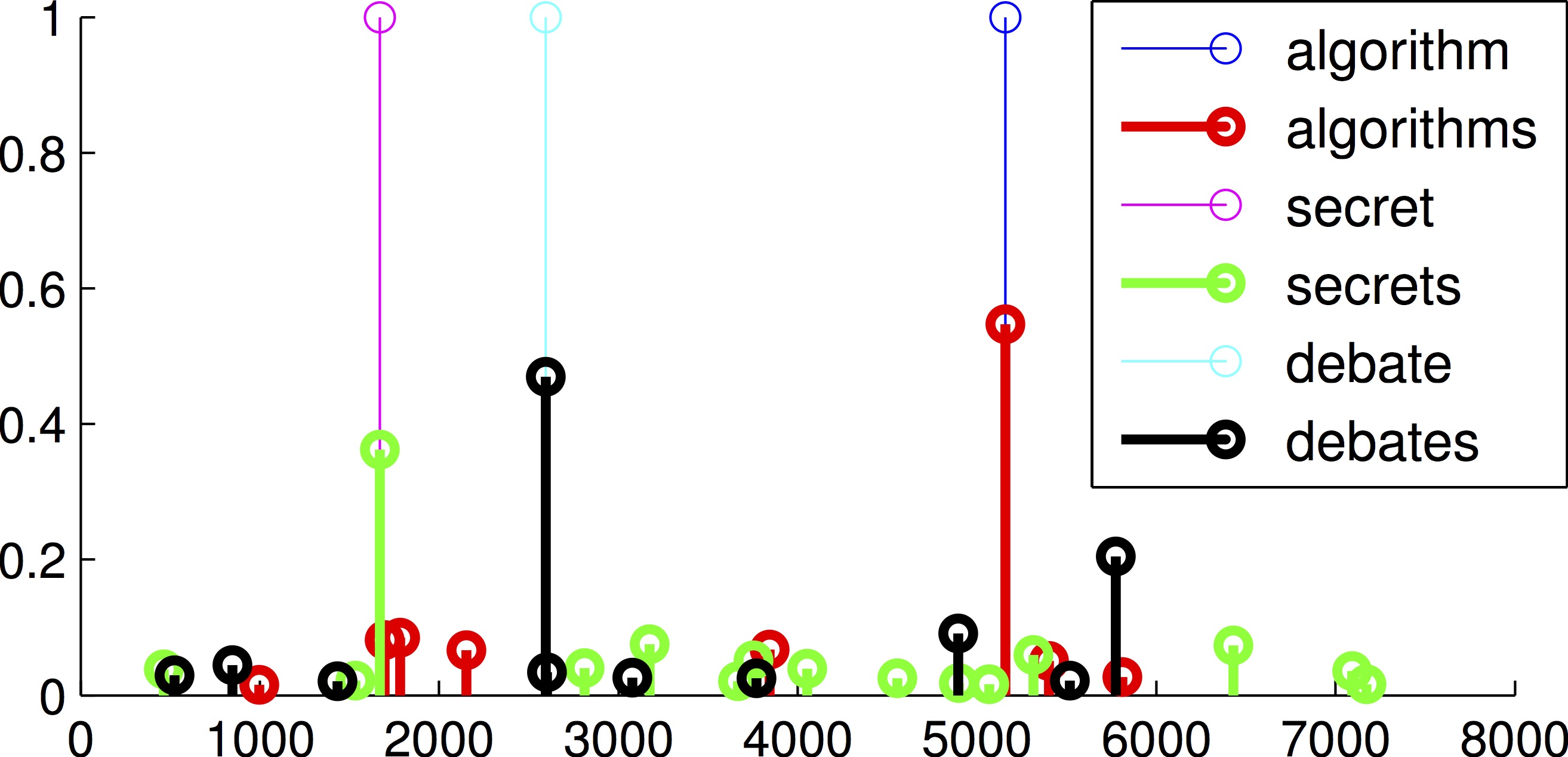}
	}
	\caption{\label{fig:sparse} The sparse representations of selected words.
	The $x$-axis is the dictionary of 8k common words; the $y$-axis is the coefficient of sparse coding. Note that \textit{algorithm}, \textit{secret}, and \textit{debate} are common words, each being coded by itself with a coefficient of 1.
	}
\end{figure}
To obtain words' sparse codes, we chose 8k common words as the ``dictionary,'' i.e., $B=8000$.
We had 2k--42k uncommon words in different settings.
We first pretrained word embeddings of both rare and common words, and obtained 200d vectors $\vector{U}$ and $\vector{w}$ in Equation~\eqref{eq:opt_sp_trick}. The dimension was specified in advance and not tuned. As there is no analytic solution to the objective, we optimized it by Adam \cite{Kingma:2014us}, which is a gradient-based method. To filter out small coefficients around zero, we simply set a value to 0 if it is less than $0.015\cdot\max\{v\in\vector{x}\}$. $w_\alpha$ in Equation~\eqref{eq:alpha} was set to 1 because we deemed fitting loss and sparsity penalty are equally important. We set $w_\beta$ in Equation~\eqref{eq:beta} to 0.1, and this hyperparameter is insensitive.

Figure~\ref{fig:sparse} plots the sparse codes of a few selected words.  As we see,  \textit{algorithm}, \textit{secret}, and \textit{debate} are common words, and each is (sparsely) coded by itself with a coefficient of 1. We further notice that a rare word like \textit{algorithms} has a sparse representation with only a few non-zero coefficient.

 Moreover, the coefficient in the code of \textit{algorithms}---corresponding to the base word \textit{algorithm}---is large ($\sim\!\!0.6$), showing that the words \textit{algorithm} and \textit{algorithms} are similar. Such phenomena are also observed with \textit{secret} and \textit{debate}.

The qualitative analysis demonstrates that our approach can indeed learn a sparse code of a word, and that the codes are meaningful.

\subsection{Quantitative Analysis of Compressed Language Models}\label{ss:eval.LM}
We then used the pre-computed sparse codes to compress neural LMs, which provides quantitative analysis of the learned sparse representations of words.
We take perplexity as the performance measurement of a language model, which
is defined by
\[
\mathrm{PPL} = 2^{-\frac{1}{N}\sum_{i=1}^{N}\log_2 p(w_i|h_i)}
\]
where $N$ is the number of running words in the test corpus.

\subsubsection{Settings}

We leveraged LSTM-RNN as the \Encoding{} subnet, which is a prevailing class of neural networks for language modeling \cite{Sundermeyer:2015gj,visualize}. The hidden layer was 200d. We used the Adam algorithm to train our neural models. The learning rate was chosen by validation from $\{0.001,0.002,0.004,0.006,0.008\}$. Parameters were updated with a mini-batch size of 256 words. We trained neural LMs by NCE, where we generated 50 negative samples for each positive data sample in the corpus. All our model variants and baselines were trained with the same pre-defined hyperparameters or tuned over a same candidate set; thus our comparison is fair. 

We list our compressed LMs and competing methods as follows.
\begin{compactitem}[\ \ \ $\bullet$]
\item \textbf{KN3}. We adopted the modified Kneser-Ney smoothing technique to train a 3-gram LM; we used the SRILM toolkit \cite{toolkit} in out experiment.
\item \textbf{LBL5}. A \textbf{L}og-\textbf{B}i\textbf{L}inear model introduced in \newcite{three}. We used 5 preceding words as context.
\item \textbf{LSTM-s}. A standard LSTM-RNN language model which is applied in \newcite{Sundermeyer:2015gj} and \newcite{visualize}. We implemented the LM ourselves based on Theano \cite{2016arXiv160502688short} and also used NCE for training.
\item \textbf{LSTM-z}. An LSTM-RNN enhanced with the \texttt{ZRegression} mechanism described in Section~\ref{ss:NCE}.
\item \textbf{LSTM-z$,$wb}. Based on LSTM-z, we compressed word embeddings in \Embedding{} and the output weights and biases in \Prediction{}.
\item \textbf{LSTM-z$,$w}. In this variant, we did not compress the bias term in the output layer. For each word in $\mathcal{C}$, we assigned an independent bias parameter.
\end{compactitem}

\subsubsection{Performance}

Tables~\ref{tab:ppl} shows the perplexity of our compressed model and baselines.
As we see, LSTM-based LMs significantly outperform the log-bilinear model as well as the backoff 3-gram LM, even if the 3-gram LM is trained on a much larger corpus with 1.6 billion words.
The {\tt ZRegression} mechanism improves the performance of LSTM to a large extent, which is unexpected. Subsection~\ref{ss:eval.zregression} will provide more in-depth analysis.

\begin{table}[!t]
\centering
\resizebox{.4\textwidth}{!}{
\begin{tabular*}{0.48\textwidth}{@{\extracolsep{\fill}}lr@{\extracolsep{0pt}}@{.}lccc}
\toprule
Vocabulary           &  \multicolumn{2}{c}{10k}    & 22k    & 36k   & 50k    \\
\midrule
KN3$^\dag$           &   90&4  & 125.3  & 146.4 & 159.9  \\
LBL5                 &  116&6  & 167.0  & 199.5 & 220.3  \\
LSTM-s               &  107&3  & 159.5  & 189.4 & 222.1  \\
LSTM-z               &   75&1  & 104.4  & 119.6 & 130.6  \\
LSTM-z$,$wb          &   73&7  & 103.4  & 122.9 & 138.2  \\
LSTM-z$,$w           &   72&9  & 101.9  & 119.3 & 129.2  \\
\bottomrule
\end{tabular*}
}
\caption{Perplexity of our compressed language models and baselines. $^\dag$Trained with the full corpus of 1.6 billion running words.}\label{tab:ppl}
\end{table}

\begin{table}[!t]
\centering
\resizebox{.4\textwidth}{!}{
\begin{tabular*}{0.48\textwidth}{@{\extracolsep{\fill}}lcccc}
\toprule
Vocabulary   &  10k    &  22k    & 36k   &  50k   \\
\midrule
LSTM-z,w     &  17.76  &  59.28  & 73.42 & 79.75  \\
LSTM-z,wb    &  17.80  &  59.44  & 73.61 & 79.95  \\
\bottomrule
\end{tabular*}
}
\caption{Memory reduction (\%) by our proposed methods in comparison with the uncompressed model LSTM-z. The memory of sparse codes are included.\label{tab:para}}
\end{table}
\begin{figure}[!t]
\resizebox{0.48\textwidth}{!}{
  \includegraphics{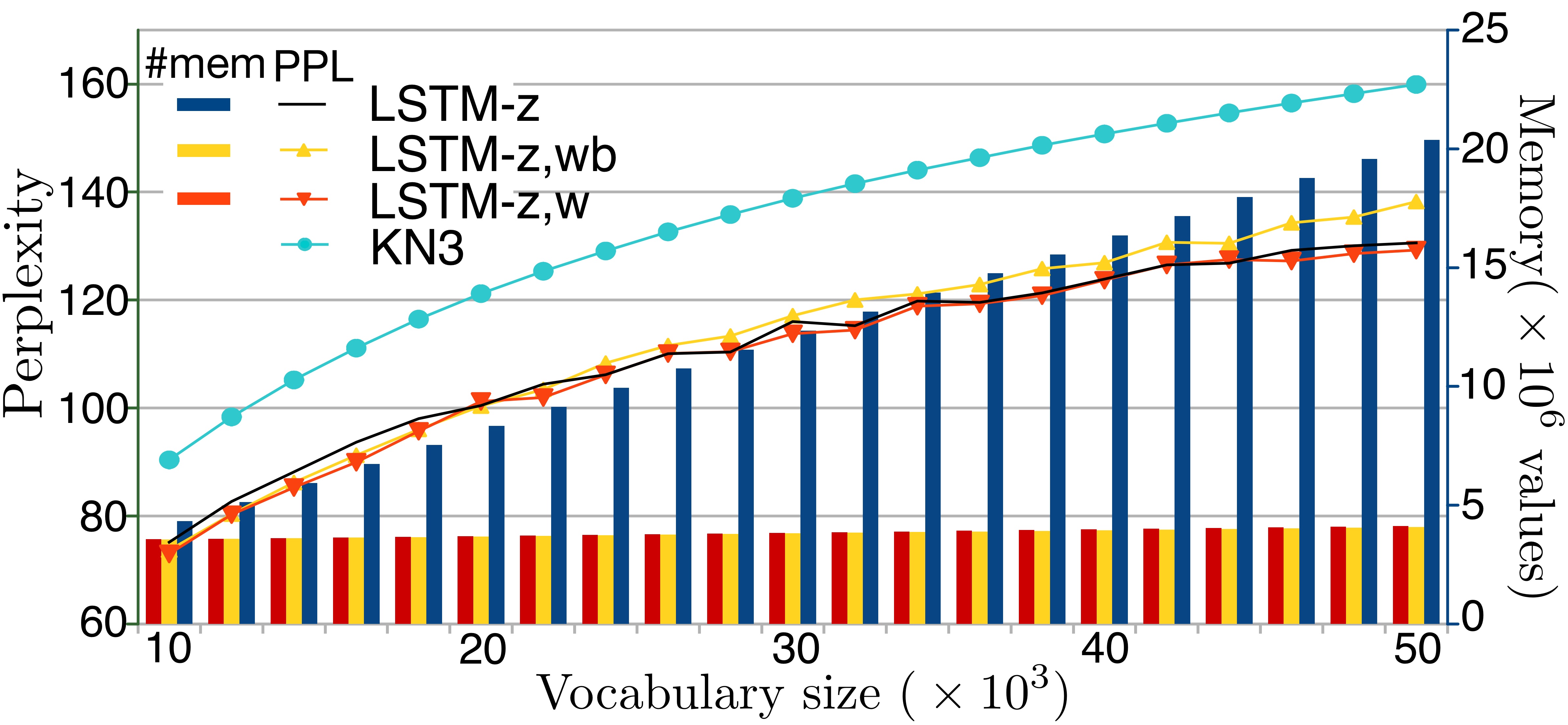}}
  \caption{Fine-grained plot of performance (perplexity) and memory consumption (including sparse codes) versus the vocabulary size.\label{fig:performance}}
\end{figure}

\begin{figure*}[!t]
    \centering
    \begin{subfigure}[t]{0.33\textwidth}
    \mbox{
        \resizebox{\textwidth}{!}{
     		\includegraphics{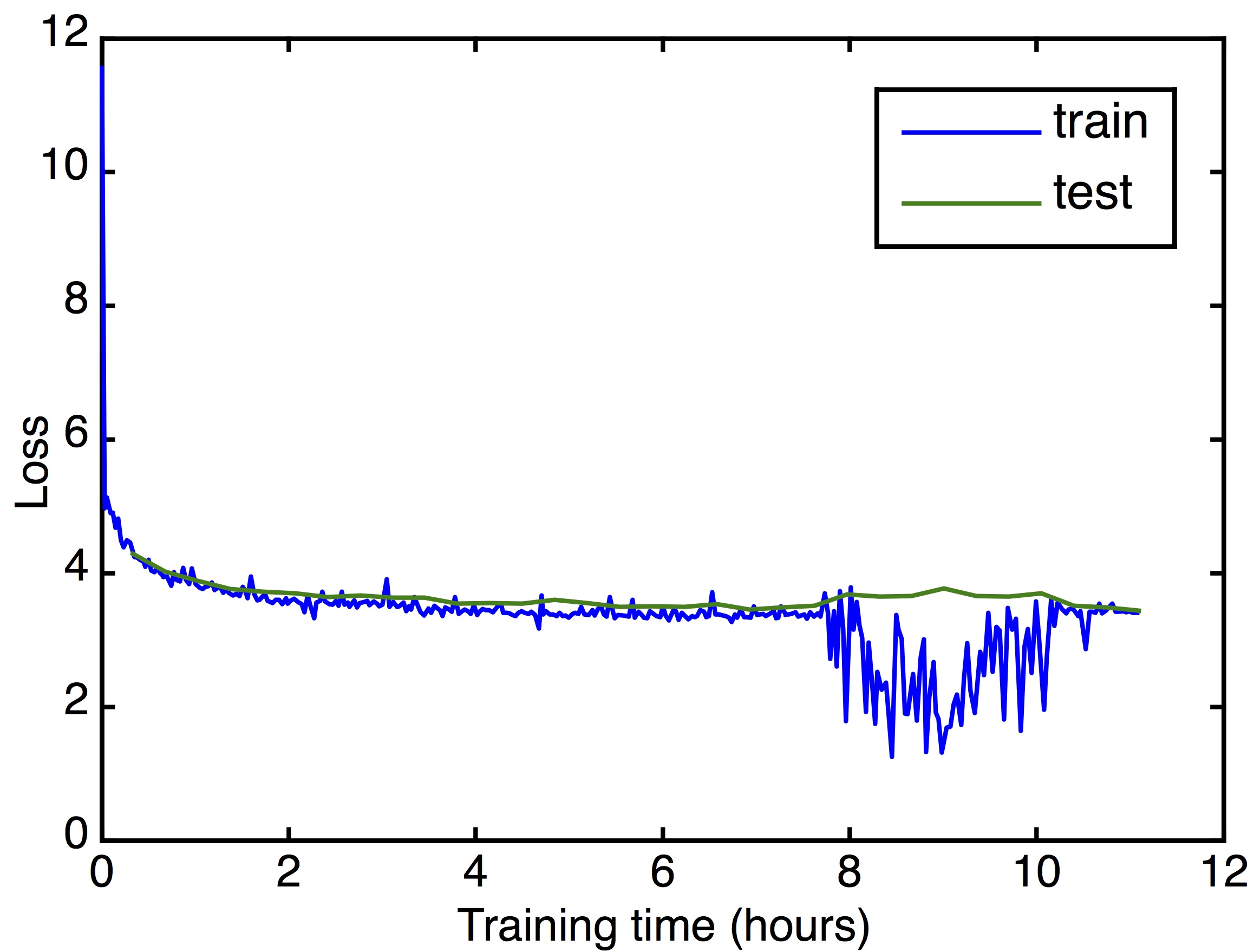}}}
  		\caption{\label{fig:loss:nce0}Training/test loss vs.~training time w/o \texttt{ZRegression}.}
    \end{subfigure}%
    \quad\ \
    \begin{subfigure}[t]{0.5\textwidth}
    \mbox{
 		\resizebox{\textwidth}{!}{
 			\includegraphics{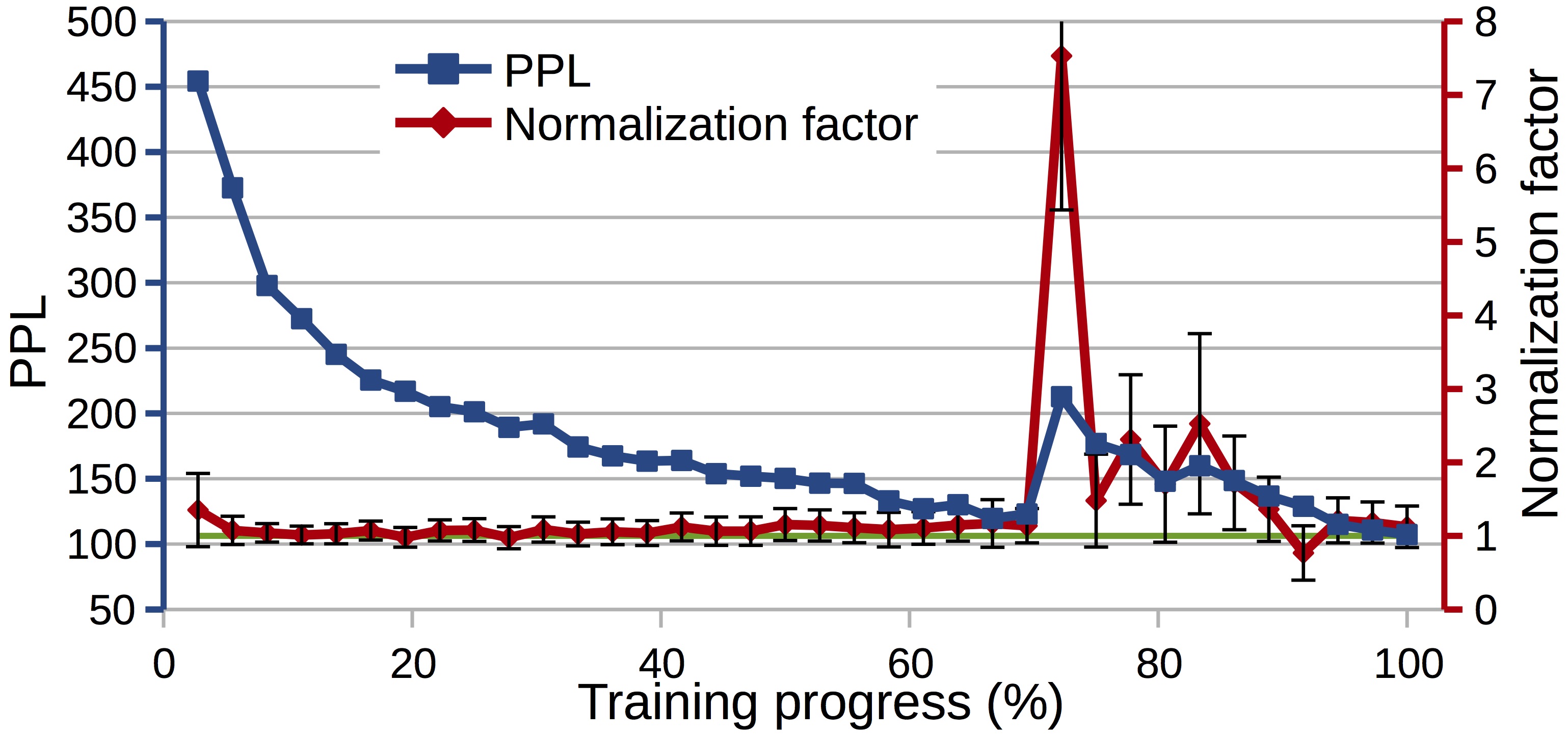}}}
 		\caption{\label{fig:sum_prob_nce0}The validation perplexity and normalization factor $Z_h$ w/o \texttt{ZRegression}. }
 	\end{subfigure}

\medskip
    \begin{subfigure}[t]{0.33\textwidth}
    \mbox{
        \resizebox{\textwidth}{!}{
     		\includegraphics{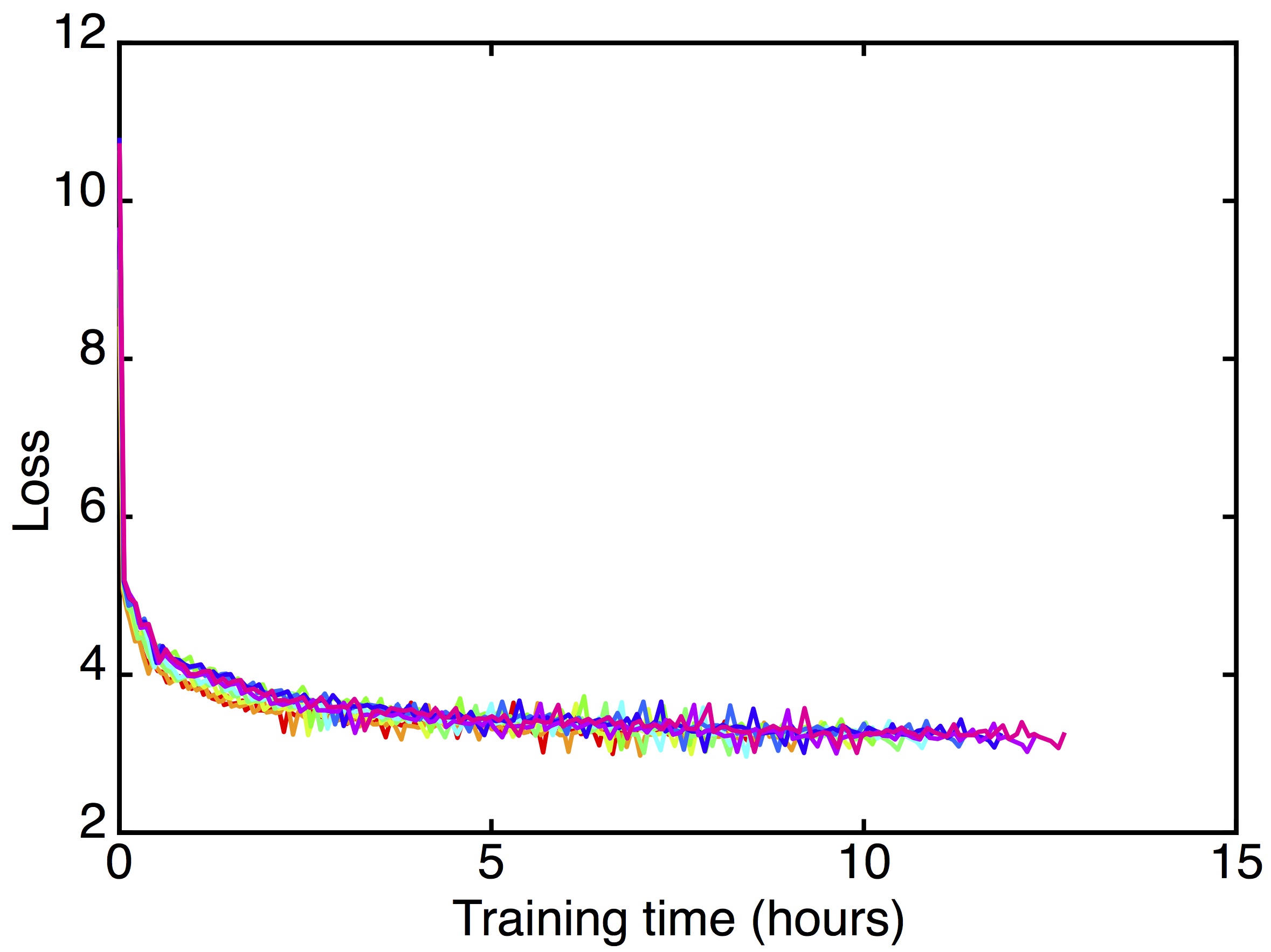}}}
 		\caption{\label{fig:loss:nce2}Training loss vs.~training time w/ \texttt{ZRegression} of different runs.}
    \end{subfigure}%
    \quad\ \
 	\begin{subfigure}[t]{0.5\textwidth}
 	\mbox{
 		\resizebox{\textwidth}{!}{
 			\includegraphics{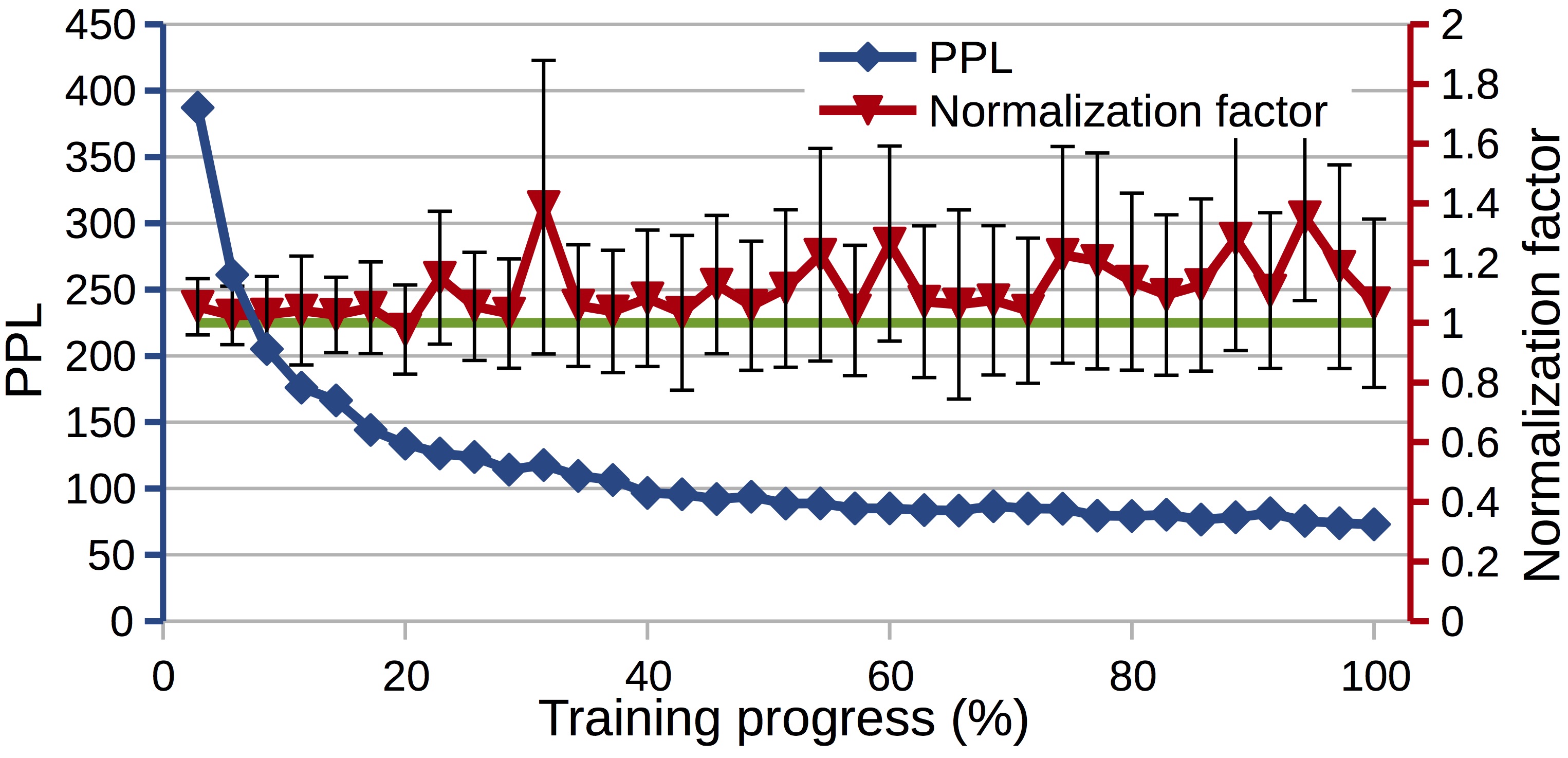}}}
 		\caption{\label{fig:sum_prob_nce2}The validation perplexity and normalization factor $Z_h$ w/ \texttt{ZRegression}.}
 	\end{subfigure}
 	\caption{Analysis of \texttt{ZRegression}.}\label{fig:zregression}
\end{figure*}

Regarding the compression method proposed in this paper, we notice that LSTM-z$,$wb and LSTM\nobreakdash-z$,$w yield similar performance to LSTM-z. In particular, LSTM-z$,$w outperforms LSTM-z in all scenarios of different vocabulary sizes. Moreover, both LSTM-z$,$wb and LSTM-z$,$w can reduce the memory consumption by up to $80\%$ (Table~\ref{tab:para}). 

We further plot in Figure~\ref{fig:performance} the model performance (lines) and memory consumption (bars) in a fine-grained granularity of vocabulary sizes.
We see such a tendency that compressed LMs (LSTM-z$,$wb and LSTM-z$,$w, yellow and red lines) are generally better than LSTM-z (black line) when we have a small vocabulary. However, LSTM-z$,$wb is slightly worse than LSTM-z if the vocabulary size is greater than, say, 20k. The LSTM-z$,$w remains comparable to LSTM-z as the vocabulary grows.

To explain this phenomenon, we may imagine that the compression using sparse codes  has two effects: it loses information, but it also enables more accurate estimation of parameters especially for rare words. When the second factor dominates, we can reasonably expect a high performance of the compressed LM.

From the bars in Figure~\ref{fig:performance}, we observe that traditional LMs have a parameter space growing linearly with the vocabulary size. But the number of parameters in our compressed models does not increase---or strictly speaking, increases at an extremely small rate---with vocabulary.

These experiments show that our method can largely reduce the parameter space with even performance improvement. The results also verify that the sparse codes induced by our model indeed capture meaningful semantics and are potentially useful for other downstream tasks.

\subsection{Effect of \texttt{ZRegression}}\label{ss:eval.zregression}

We next analyze the effect of \texttt{ZRegression} for NCE training. As shown in Figure~\ref{fig:zregression}a, the training process becomes unstable after processing 70\% of the dataset: the training loss vibrates significantly, whereas the test loss increases.

We find a strong correlation between unstableness and the $Z_h$ factor in Equation~\eqref{eq:nrm_lang_model}, i.e., the sum of unnormalized probability (Figure~\ref{fig:zregression}b). Theoretical analysis shows that the $Z_h$ factor tends to be self-normalized even though it is not forced to \cite{Gutmann:2012tr}.
However, problems would occur, should it fail.

In traditional methods, NCE jointly estimates normalization factor $Z$ and model parameters \cite{Gutmann:2012tr}. For language modeling, $Z_h$ dependents on context $h$. \newcite{Mnih:2012tv} propose to estimate a separate $Z_h$ based on two history words (analogous to 3-gram), but their approach hardly scales to RNNs because of the exponential number of different combinations of history words.

We propose the \texttt{ZRegression} mechanism in Section~\ref{ss:NCE}, which can estimate the $Z_h$ factor well (Figure~\ref{fig:zregression}d) based on the history vector $\vector{h}$. In this way, we manage to stabilize the training process (Figure~\ref{fig:zregression}c) and improve the performance by a large margin, as has shown in Table~\ref{tab:ppl}.

It should be mentioned that \texttt{ZRegression} is not specific to model compression and is generally applicable to other neural LMs trained by NCE.

\section{Conclusion}
In this paper, we proposed an approach to represent rare words by sparse linear combinations of common ones. Based on such combinations, we managed to compress an LSTM language model (LM), where memory does not increase with the vocabulary size  except a bias and a sparse code for each word. Our experimental results also show that the compressed LM has yielded a better performance than the uncompressed base LM.

\section*{Acknowledgments}
This research is supported by the National Basic Research Program of China (the 973 Program) under Grant No.~2015CB352201, the National Natural Science Foundation of China under Grant Nos.~61232015, 91318301, 61421091 and 61502014, and the China Post-Doctoral Foundation under Grant No.~2015M580927.

\bibliographystyle{acl2016}
\bibliography{acl2016}

\end{document}